\documentclass[final]{cvpr}

\usepackage{times}
\usepackage{epsfig}
\usepackage{graphicx}
\usepackage{amsmath}
\usepackage{amssymb}

\usepackage{mathrsfs}
\usepackage{bm}

\usepackage{threeparttable}
\usepackage{epsfig}
\usepackage{graphicx}
\usepackage{multirow}
\usepackage{algorithmic}
\usepackage{algorithm}
\usepackage[sort,nocompress]{cite}

\usepackage[pagebackref=true,breaklinks=true,colorlinks,bookmarks=false]{hyperref}

\def\cvprPaperID{6919} 
\def\confYear{CVPR 2021}
\begin{document}

\title{Implicit Feature Alignment: Learn to Convert Text Recognizer to Text Spotter}

\author{
Tianwei Wang,\footnotemark[1] \textsuperscript{\rm 1 \rm 2}
Yuanzhi Zhu,\footnotemark[1] \textsuperscript{\rm 1 \rm 2}
Lianwen Jin,\textsuperscript{\rm 1 \rm 3}
Dezhi Peng,\textsuperscript{\rm 1}
Zhe Li,\textsuperscript{\rm 1}\\
Mengchao He,\textsuperscript{\rm 2}
Yongpan Wang,\textsuperscript{\rm 2}
Canjie Luo\textsuperscript{\rm 1}\\
\textsuperscript{\rm 1}School of Electronic and Information Engineering, South China University of Technology. \\
\textsuperscript{\rm 2}Alibaba Group.  
\textsuperscript{\rm 3}Guangdong Artificial Intelligence and Digital Economy Laboratory. \\
\{wangtw, z.yuanzhi, zheli0205\}@foxmail.com, \{eelwjin, eedzpeng\}@scut.edu.cn,\\
mengchao.hmc@alibaba-inc.com, yongpan@taobao.com, canjie.luo@gmail.com.
}

\maketitle
\thispagestyle{empty}
\renewcommand{\thefootnote}{\fnsymbol{footnote}} 
\footnotetext[1]{Equal contribution. This work is done during Tianwei Wang and Yuanzhi Zhu's internship at Alibaba Group. Lianwen Jin is the corresponding author.} 
\begin{abstract}
Text recognition is a popular research subject with many associated challenges.
Despite the considerable progress made in recent years, the text recognition task itself is still constrained to solve the problem of reading cropped line text images and serves as a subtask of optical character recognition (OCR) systems.
As a result, the final text recognition result is limited by the performance of the text detector. 
In this paper, we propose a simple, elegant and effective paradigm called Implicit Feature Alignment (IFA), which can be easily integrated into current text recognizers, resulting in a novel inference mechanism called IFA-inference.
This enables an ordinary text recognizer to process multi-line text such that text detection can be completely freed.
Specifically, we integrate IFA into the two most prevailing text recognition streams (attention-based and CTC-based) and propose attention-guided dense prediction (ADP) and Extended CTC (ExCTC). 
Furthermore, the Wasserstein-based Hollow Aggregation Cross-Entropy (WH-ACE) is proposed to suppress negative predictions to assist in training ADP and ExCTC. 
We experimentally demonstrate that IFA achieves state-of-the-art performance on end-to-end document recognition tasks while maintaining the fastest speed,
and ADP and ExCTC complement each other on the perspective of different application scenarios.
Code will be available at https://github.com/Wang-Tianwei/Implicit-feature-alignment.
\end{abstract}

\section{Introduction}
Text is extensively used in people's daily life, delivering rich and useful visual information.
Reading text in images is one of the most important tasks in the field of computer vision.

\begin{figure}[t]
\centering
\includegraphics[width=0.45\textwidth]{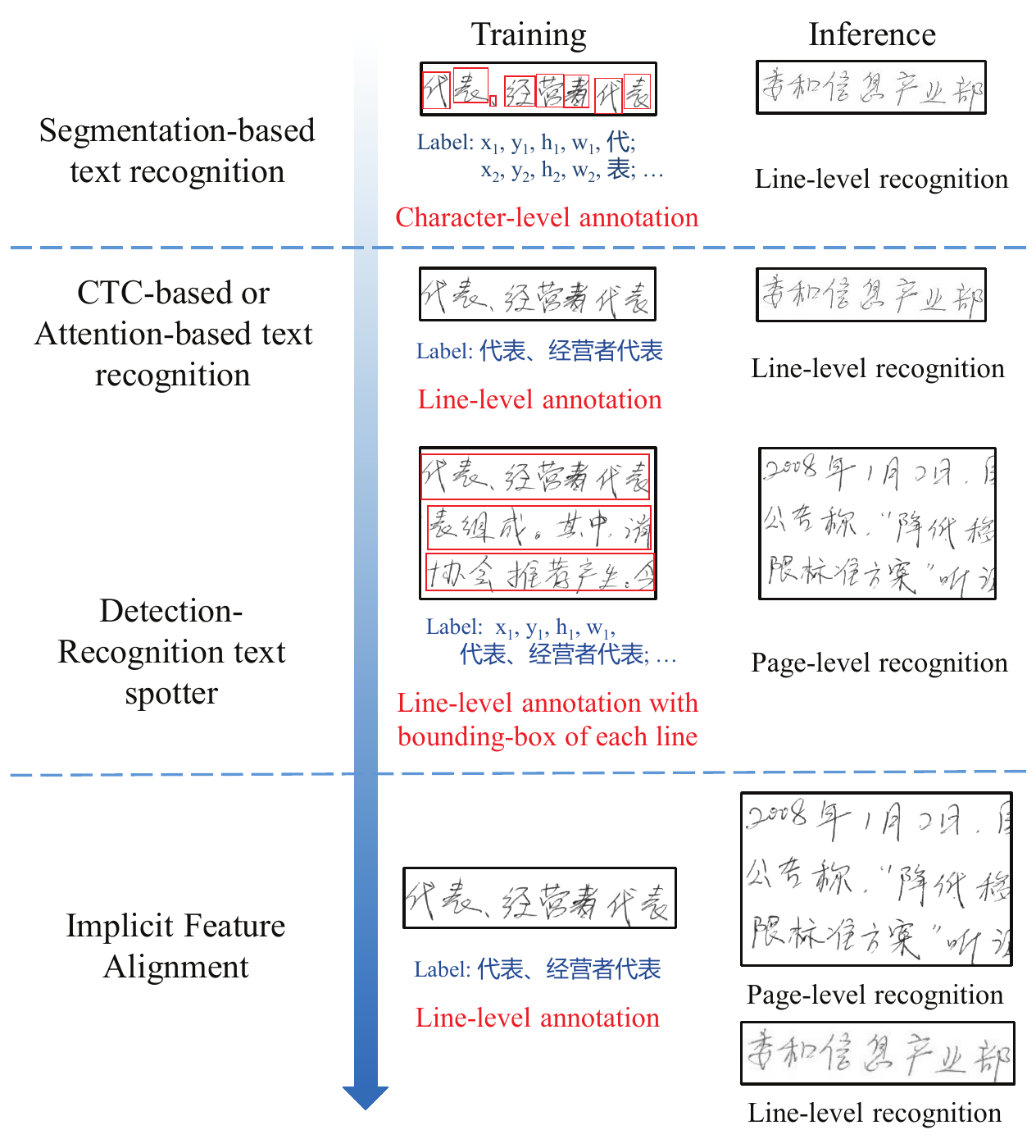}
\caption{The development of text recognizers.}
\label{Figure_fig1_cmp}
\end{figure}

Most OCR systems follow the pipeline that first uses a text detector to detect the location of each text line and then recognizes the detected line texts with a text recognizer.
Under this pipeline, the performance of the entire system is determined by the cascading of each module, and the performance degradation of each module leads to the deterioration of the overall performance.
Although many end-to-end (E2E) OCR methods \cite{2017Deep,lyu2018mask,li2017towards,wang2020all,qin2019towards,liu2020abcnet,qiao2019text,feng2019textdragon} have been proposed in recent years, they have still used such a pipeline, and considerable efforts have been made to better develop the bridge between text detectors and text recognizers. 
Thus, the error accumulation problem has not been solved.
To remedy this issue, a direct way is shortening this pipeline by extending text recognizer from text-line to full-page level.
In this paper, we propose a simple, yet effective, paradigm, called \textbf{I}mplicit \textbf{F}eature \textbf{A}lignment (IFA), to realize detection-free full-page text recognition with state-of-the-art performance and significantly faster inference speed as compared to previous models.

Alignment is the core issue in the design of a text recognizer. This means the way to teach the recognizer the location of each character, as well as the category to which it belongs.
As shown in Fig.~\ref{Figure_fig1_cmp}, the development of text recognition methods has shown a trend of \textbf{more general recognition with fewer annotations}.
In the early research, integrated segmentation-recognition methods \cite{wang2012handwritten,zhou2013handwritten,zhou2014minimum,wu2017improving} construct the segmentation-recognition lattice based on sequential character segments of line text images, followed by optimal path searching by integrating the recognition scores, geometry information, and semantic context.
These methods require text annotations with a bounding box for each character to train a line-level recognizer.
With the rapid development of deep learning technology, Connectionist Temporal Classification (CTC) \cite{Graves2006Connectionist,graves2009novel} and the attention mechanism \cite{bahdanau2015neural,shi2018aster,lee2016recursive,shi2016robust} can realize training of text recognizers with text annotations.
More specifically, CTC assumes that the character order in the label and that in the image are monotonous, and thereby designs a series of many-to-one rules to align the dense frame-wise output with the label.
The attention mechanism uses a parameterized attention decoder to align and transcribe each character on the image.
However, both CTC and the attention mechanism have their own limitations, which makes them impossible to conduct full-page recognition.
CTC can only align two 1-D sequences, ignoring the height information. Thus, it cannot process multi-line images. 
Attention relies on a parameterized attention module, whose performance decays as the sequence length increases \cite{DAN_aaai20,cong2019comparative}, not to mention the full-page level.

The proposed IFA aims to overcome the aforementioned limitations and convert a text recognizer trained on line text images to recognize multi-line texts.
IFA leverages the learnable aligning ability of the deep learning model to automatically align and train positive pixels (\ie, pixels on a feature map that contains character information), thereby realizing conversion from a text recognizer to a text spotter.
We integrate IFA with CTC and the attention mechanism, producing two new methods called attention-guided dense prediction (ADP) and extended CTC (ExCTC).
For ADP, we theoretically analyze the optimization objective of the attention mechanism at each step and find it equivalent to ensuring the aligned pixels being correctly recognized.
For ExCTC, we add a pluggable squeeze module that  learns to align the positive pixels on each column to conduct feature collapse.
In addition to aligning positive pixels by ADP and ExCTC, we modify Aggregation Cross-Entropy (ACE) \cite{Xie2019Aggregation} and propose Wasserstein-based Hollow ACE (WH-ACE) to suppress negative noise. 
Both ADP and ExCTC have a unified inference form called IFA-inference, which can process single-line, multi-line, or even full-page text recognition tasks.

\section{Related Work}
\subsection{Text Recognition}
Based on the decoding strategy, there are two main categories of methods for text recognition. 
The first form can be be collectively called \textbf{dense prediction}, where the classification is performed in a per-pixel prediction manner; that is, each pixel in the feature map is interpreted as a probability distribution. 
The other is \textbf{attention-based}, where an attention decoder is adopted to automatically align the feature with the corresponding ground-truth character in a sequential manner.

More specifically, dense prediction contains several implementations. 
CTC \cite{Graves2006Connectionist,graves2009novel} outputs a dense probability distribution frame by frame for 1-D monotonic sequence, and designs a series of rules to align the dense output with the label. 
It essentially solves the problem of 1-D sequence alignment without character-level annotation. 
Aggregation cross entropy (ACE) \cite{Xie2019Aggregation} attempts to convert the problem of sequence recognition to the alignment of two aggregated probabilities. It involves a strong assumption; that is, the characters with the same number of occurrences in the label have equal distribution probability.
Generally, some segmentation-based methods \cite{liao2019scene,wan2020textscanner} are proposed to handle irregular scene text recognition task, this type can also be regarded as an implementation of dense prediction.
Attention-based methods have variants of forms \cite{bahdanau2015neural,luong2015effective,li2019show,DAN_aaai20,yang2017learning}, but the fundamental idea of using an attention map to align each character is unchanged.
With a learnable alignment module, attention-based methods are more flexible when facing different application scenarios. 
Without a heavily parameterized decoder, dense prediction methods are usually significantly faster than attention-based methods.

Despite the considerable progress made in curved text rectificationn \cite{shi2018aster,cluo2019moran,liao2019scene,liu2018char,zhan2019esir,yang2019symmetry}, data generation \cite{jaderberg2014synthetic,gupta2016synthetic,zhan2019ga,long2020unrealtext,zhan2018verisimilar}, and decoder design \cite{li2019show,DAN_aaai20,zhu2019text,Deli2020Towards,wang2019scene,qiao2020seed,litman2020scatter}, the text recognition task itself remains constrained to solve the problem of reading cropped line text images and serves as a subtask of full-page text recognition (also known as text spotting).

\subsection{Full-Page Recognition (Text Spotting)}
End-to-end text spotting has evolved as a new trend in the past few years. 
Most existing methods cascade text detector and text recognizer through pooling-like operations.
Michal \emph{et al.} \cite{2017Deep} proposes deep text spotter, which adopts YOLOv2 as text detector and a CTC based recognizer. A bilinear sampler serves as the bridge between these two modules.
Li \emph{et al.} \cite{li2017towards} proposes to cascade Faster-RCNN \cite{ren2015faster} and attention decoder through RoIPooling to realize end-to-end spotting.
From then on, lots of novel methods are proposed \cite{qin2019towards,lyu2018mask,liu2020abcnet}.
Most of these methods concentrate on how to bridge text detectors and text recognizers in a better manner and introduce methods such as RoISlide \cite{feng2019textdragon}, RoI masking \cite{qin2019towards}, character region attention \cite{baek2020character}, and BezierAlign\cite{liu2020abcnet}.

On document OCR tasks, Curtis \emph{et al.} \cite{2018Start} propose to find the start position of each line and then incrementally follow and read line texts.
Bluche \cite{bluche2016joint} proposes the use of a sequence decoder with weighted collapse to directly recognize multi-line images.
Mohamed \emph{et al.} \cite{yousef2020origaminet} proposes to implicitly unfold a multi-line image into a single line image and use CTC to supervise the output.

Different from these methods, IFA attempts to enable an ordinary text recognizer trained with only textline data to be able to process page-level text recognition. 

\section{Methodology}

IFA-inference can be modeled as $\bm{Y} = \mathbb{C}(\mathbb{F}(\bm{x}))$, where a CNN based feature extractor $\mathbb{F}$ encodes the input image $\bm{x}$ with label $\bm{s}=\{s_1, s_2, \cdots, s_T \}$ into feature map $\bm{F}$:
\begin{align}
\bm{F}=\mathbb{F}(\bm{x}), \bm{F}\in\mathbb{R}^{C \times H \times W}, 
\end{align}
where $C$ , $H$ and $W$ denote the output channels, height and width of the feature map, respectively. 
Then, the classifier $\mathbb{C}$ with trainable parameter $W_c$ classifies each pixel in the feature map as:
\begin{align}
\bm{Y}=\mathbb{C}(\bm{F}), \bm{Y}\in\mathbb{R}^{K \times H \times W},
\end{align}
where $K$ denotes the total number of character classes plus a blank symbol.

\begin{figure}[t]
\centering
\includegraphics[width=0.4\textwidth]{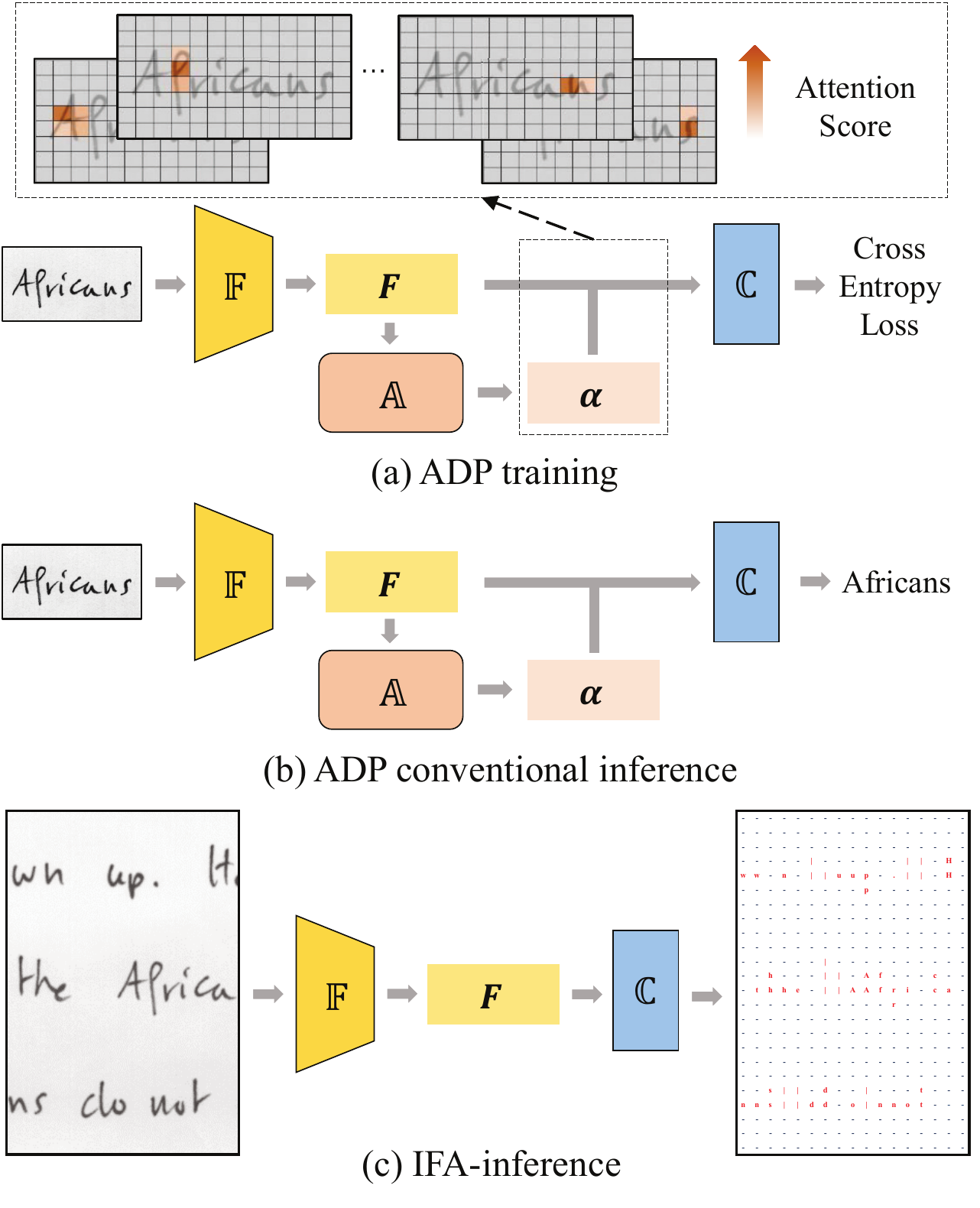}
\caption{ADP. (a) ADP training. (b) conventional attention inference. (c) IFA-inference derived from ADP.}
\label{Figure_ADP}
\end{figure}

\subsection{Attention-guided Dense Prediction (ADP)}

Generally, an attention decoder (represented as $\mathbb{A}$) outputs text sequentially. 
In this section, we theoretically analyze the optimization objective of the attention mechanism at each step and explain the derivation procedure of the IFA inference.

\subsubsection{A general form of the attention mechanism.}

Assuming that the attention map at decoding step $t$ is $\alpha_t$, which is yielded by softmax function
\begin{align}
\label{eq_alpha}
\alpha_{t,h,w} = softmax (e_{t,h,w}).
\end{align}
Here, $e_{t,h,w}$ is a score map generated by score function (for details please refer to \cite{bahdanau2015neural,luong2015effective}). 
Then, we can calculate the context vector $c_t$ by applying the attention map over $\bm{F}$ with the Hadamard product
\begin{align}
\label{eq_ct}
c_t = \sum_{H,W} \alpha_{t,h,w} F_{h,w}.
\end{align}
Now, $c_t$ is classified by $\mathbb{C}$ as
\begin{align}
\label{eq_wct}
y_t = \mathbb{C}(c_t) = W_c c_t.
\end{align}
Finally, the loss at time $t$ can be calculated as
\begin{align}
\label{eq_loss}
L_t = -log(softmax(y_{t})_{s_t}).
\end{align}
This maximizes the probability of generating character $s_t$ at this step.
Fig.~\ref{Figure_ADP} (a) illustrates the training stage of ADP, where the attention decoder at each step highlights the corresponding positive pixels.

\subsubsection{IFA in the attention mechanism.}

In Eq.~\ref{eq_loss}, the optimization objective at step $t$ is to maximize $softmax(y_{t})_{s_t}$.
According to Eqs.~\ref{eq_ct} and \ref{eq_wct}, we have:
\begin{align}
y_t = W_c c_t = W_c \sum_{H,W} \alpha_{t,h,w} F_{h,w} = \sum_{H,W} \alpha_{t,h,w} W_c F_{h,w}.
\end{align}
As shown in Eq.~\ref{eq_alpha}, the attention map is \textbf{nearly one-hot} because of the softmax function. 
Thus, we have
\begin{align}
y_t = \sum_{H,W} \alpha_{t,h,w} W_c F_{h,w} \approx \alpha_{t,h',w'} W_c F_{h',w'} \approx W_c F_{h',w'},
\end{align}
where $h', w' = argmax(\alpha_{t,h,w})$, representing the attention center at step $t$. 
Therefore, the optimization objective at step $t$ can be approximately considered to maximize $softmax(W_c F_{h',w'})_{s_t}$; that is, \textbf{ensuring that the aligned pixels are correctly recognized}. 

Figs.~\ref{Figure_ADP} (b) and (c) describe the conversion from conventional attention inference to IFA-inference by removing the attention decoder.
As $\mathbb{C}$ has been trained to recognize positive pixels, after removing the attention module, the aligned positive pixels can still be recognized.

\subsection{Extended CTC (ExCTC)}
Unlike the attention mechanism which aligns each character during training, the CTC algorithm aligns two sequences.
To allow CTC training, we use a column squeeze module to squeeze the 2-D feature map into 1-D feature sequence.
This module highlights the key pixels of each column by conducting an attention operation in each column.

\subsubsection{Integrate IFA into CTC: Column squeeze.}

The column squeeze module $\mathbb{S}$ is used to squeeze 2-D feature map $\bm{F}\in\mathbb{R}^{C \times H \times W}$ into 1-D feature sequence $\bm{F'}\in\mathbb{R}^{C \times 1 \times W}$ along the vertical direction using a column attention map.
$\mathbb{S}$ consists of two stacked convolutional layers with batch normalization and ReLU function.
Similar modules have been used in \cite{bluche2016joint} and \cite{huang2019attention}.
The vertical attention map $\bm{\alpha}$ is calculated as
\begin{align}
\bm{e} = \mathbb{S}(\bm{F}),
\end{align}
\begin{align}
\alpha_{h,w}=\frac{exp(e_{h,w})}{\sum_{h'=1}^{H} exp(e_{h',w})}.
\end{align}
Then, $\bm{F'}$ is calculated as
\begin{align}
F'_w = \sum_{h=1}^{H} \alpha_{h,w} F_{h,w}.
\end{align}

The following steps and training strategies are similar to those of any other conventional CTC-based methods.
First, the probability distribution $\bm{y'}$ is generated by applying $\mathbb{C}$ to $\bm{F'}$:
\begin{align}
\bm{y'} = \mathbb{C}(\bm{F'}).
\end{align}
Given CTC path $\pi$, its probability can be calculated as
\begin{align}
P(\pi|\bm{y'}) = \prod_{w=1}^{W} y_{\pi_t}^{'w}.
\end{align}
The conditional probability of the target sequence is the summation of the probabilities of all possible paths
\begin{align}
P(\bm{s}|\bm{y'}) = \sum_{\pi:\mathbb{B}(\pi)=\bm{s}} P(\pi|\bm{y'}),
\end{align}
where $\mathbb{B}$ denotes the mapping function \cite{Graves2006Connectionist}.
The loss function is negative log-likelihood of the target sequence:
\begin{align}
L_{CTC} = -log(P(\bm{s}|\bm{y'})).
\end{align}

\begin{figure}[t]
\centering
\includegraphics[width=0.4\textwidth]{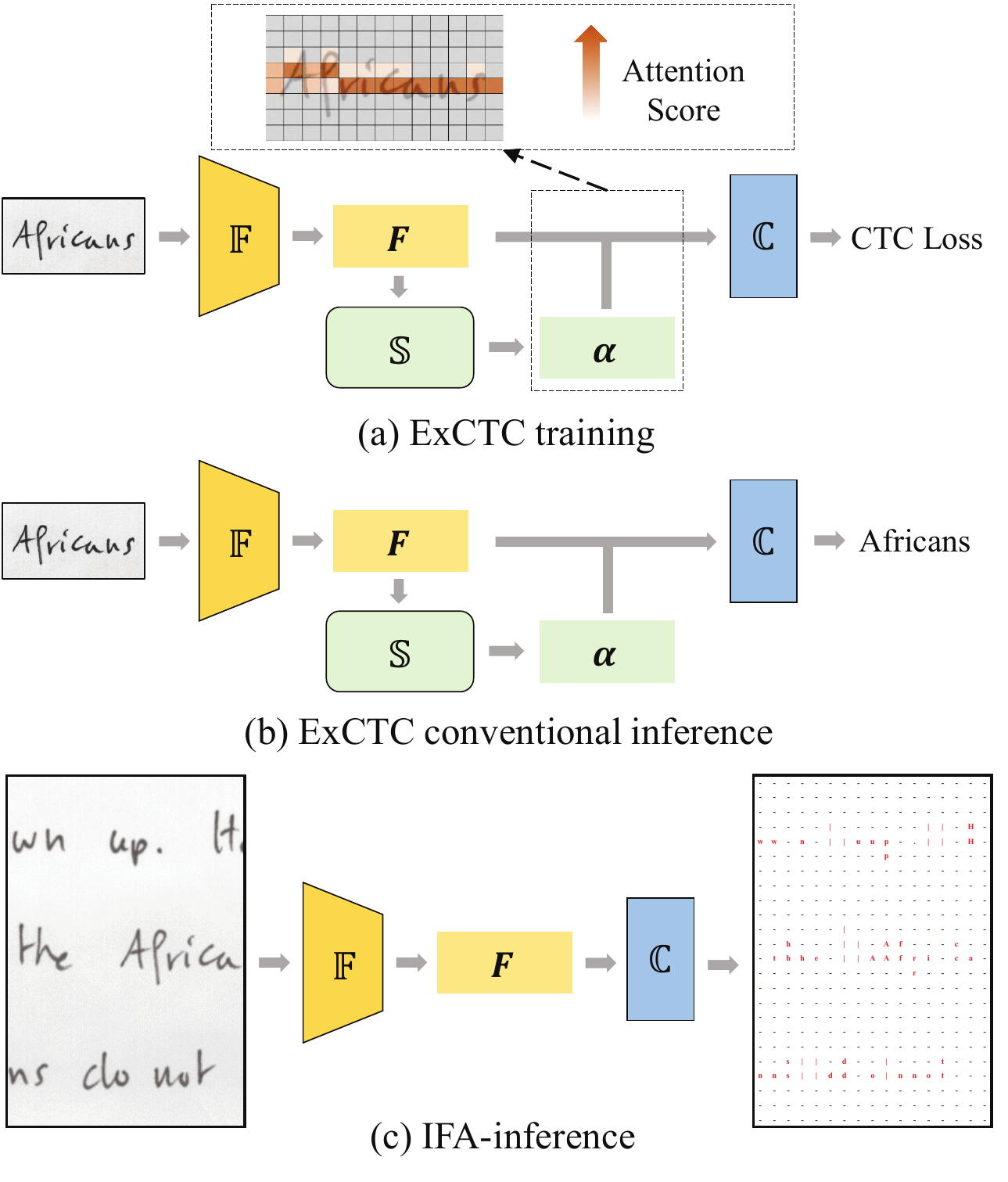}
\caption{ExCTC. (a) ExCTC training. (b) ExCTC conventional inference. (c) IFA-inference derived from ExCTC.}
\label{Figure_ExCTC}
\end{figure}

Similar to ADP, ExCTC ensures that the aligned key pixels of each column are correctly recognized.
As $\mathbb{C}$ has been trained to have the ability to recognize positive pixels on each column, after removing $\mathbb{S}$, the aligned positive pixels can still be recognized.
Fig.~\ref{Figure_ExCTC} (b) and (c) describe the conversion from conventional CTC inference to IFA-inference by removing the column squeeze module.

\begin{figure}[t]
\centering
\includegraphics[width=0.4\textwidth]{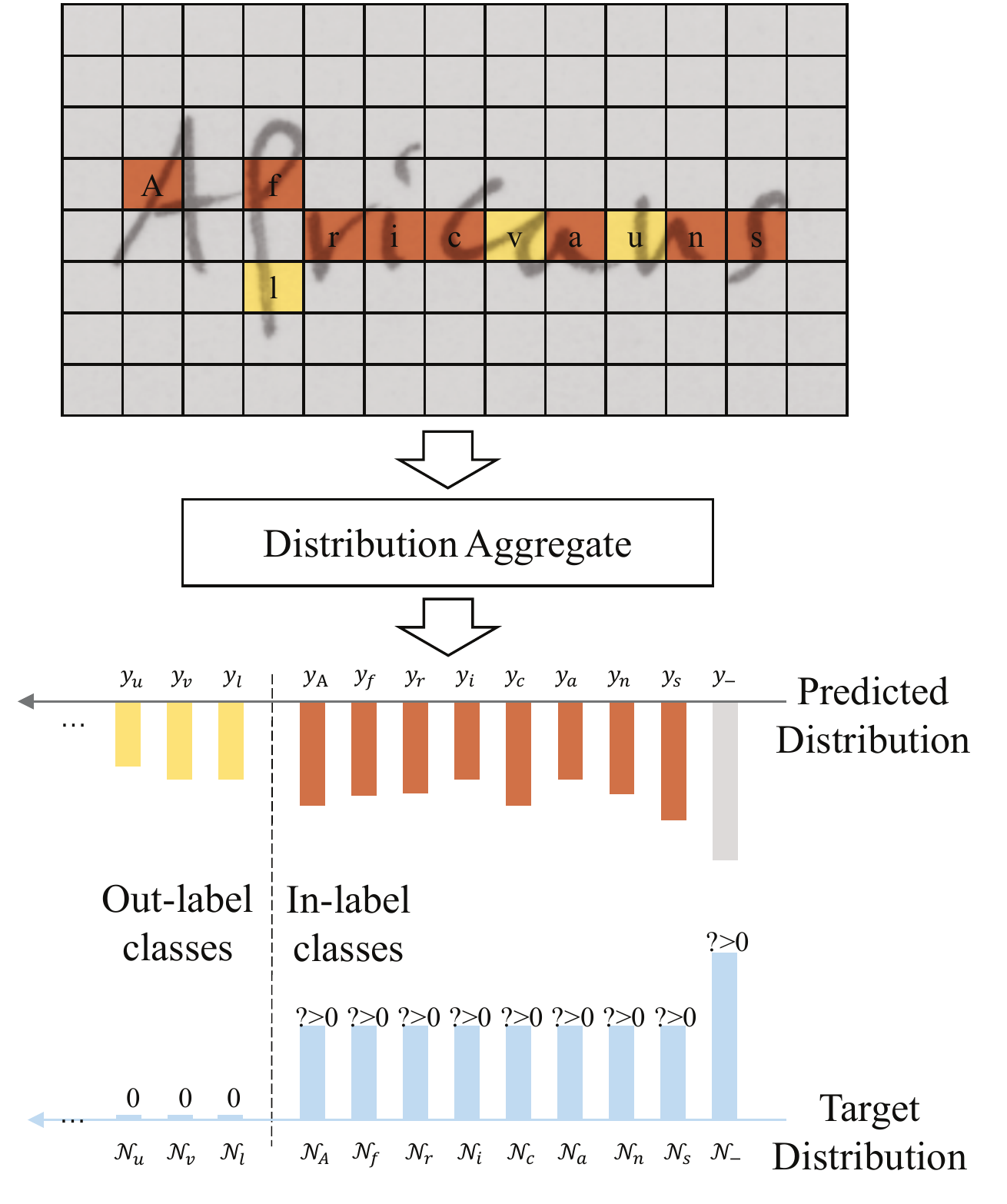}
\caption{Wasserstein-based Hollow ACE (WH-ACE). The noise suppression task is converted into a distribution optimization task.}
\label{Figure_WHACE}
\end{figure}

\subsection{Wasserstein-based Hollow ACE (WH-ACE)}
Both ADP and ExCTC simply align and train the positive pixels of $\bm{F}$, ignoring the suppression of negative predictions.
As shown in Fig.~\ref{Figure_WHACE}, although ADP and ExCTC correctly predict positive pixels, such as ``A", ``f" and ``r", they also include additional negative predictions such as ``l", ``v" and ``u". 
We define those classes included in the label as ``in-label classes" and those not included as ``out-label classes".  
In this section, we propose WH-ACE to suppress the out-label negative predictions while keeping the positive predictions unaffected.

ACE \cite{Xie2019Aggregation} is proposed to address the sequence recognition problem by comparing the predicted and target distributions.
The original version of ACE makes a strong assumption that characters with the same frequency of occurrences have the same predicted probability. 
This ignores the scale problem and, thus, necessitates an additional character number counting module to help in training.

Based on the dense prediction $\bm{Y} = \mathbb{C}(\mathbb{F}(\bm{x}))$, we denote the aggregated prediction probability as $\bm{y} = \sum_{H,W} \bm{Y} = \{y_1, y_2,...,y_K\}$ and the target distribution as $\bm{\mathcal{N}} = \{\mathcal{N}_1, \mathcal{N}_2, ..., \mathcal{N}_K\}$.

\subsubsection{Hollow parameter} 

Without character-level annotation, the concrete target distribution is unavailable.
This is because we do not know the location of each character and the number of pixels that it should contain. 
Concrete priori knowledge is only available for the in-label class. 
Its probability is greater than zero, and for the out-label class, the probability should be zero. 
This observation can be formulated as:

\begin{align}
\left\{ \begin{array}{l} 
\mathcal{N}_k > 0, \ if \ k \in \bm{s} 
\\ 
\mathcal{N}_k = 0, \ if \ k \notin \bm{s} 
\end{array} \right. 
.
\end{align}

As we only have the concrete distribution of the out-label classes, we propose a hollow parameter to correct the distribution:
\begin{align}
h_k = \left\{ \begin{array}{l} 0, \ if \ k \in \bm{s} \\ 1, \ if \ k \notin \bm{s} \end{array} \right. .
\end{align}
With the hollow parameter, the hollowed distributions can be computed as:
\begin{align}
\overline{\bm{y}} = \{h_1 y_1, h_2 y_2, ..., h_K y_K\},
\\
\bm{\mathcal{\overline{N}}} = \{h_1 \mathcal{N}_1, h_2 \mathcal{N}_2, ... ,h_K \mathcal{N}_K\}.
\end{align}
Through further analysis we can find that:
\begin{align}
\mathcal{\overline{N}}_k = 
\left\{ \begin{array}{l} 
h_k \mathcal{N}_k = 0 \times \mathcal{N}_k = 0, \ if \ k \in \bm{s} 
\\ 
h_k \mathcal{N}_k = h_k \times 0 = 0, \ if \ k \notin \bm{s} 
\end{array} \right. 
.
\end{align}
Hence, the hollowed target distribution $\bm{\overline{\mathcal{N}}} = \bm{0}$.

\subsubsection{Wasserstein distance for optimization}

It can be observed that $\overline{\bm{y}}$ and $\bm{\overline{\mathcal{N}}}$ have no overlap as all pixels in $\bm{\overline{\mathcal{N}}}$ are zero.
In \cite{Xie2019Aggregation}, the authors propose the use of cross entropy loss to compare these two probability distributions.
However, cross entropy can not optimize two non-overlap distribution, which make it not suitable in this situation.

Here we propose the use of the 1-Wasserstein distance (denote as $W_1$) to compare $\overline{\bm{y}}$ and $\bm{\overline{\mathcal{N}}}$, the loss function of WH-ACE is written as:
\begin{align}
&L_{WH-ACE} = \frac{1}{K} W_1(\overline{\bm{y}}, \bm{\overline{\mathcal{N}}}) \notag \\
&= \frac{1}{K} \sum_{k=1}^{K}|\overline{y}_k - \overline{\mathcal{N}}_k| 
= \frac{1}{K} \sum_{k=1}^{K}|h_k y_k - h_k \mathcal{N}_k| \notag \\
&= \frac{1}{K} \sum_{k=1}^{K}|h_k y_k|
= \frac{1}{K} \bm{h} \cdot \bm{y}.
\end{align}
The physical interpretation of WH-ACE is moving out-label probabilities into in-label classes.

\subsection{Training and Inference}
Combining ADP with WH-ACE, the loss function of ADP training is
\begin{align}
L_{ADP} = \sum_{t=1}^{T}{L_t} + L_{WH-ACE}.
\end{align}
Combining ExCTC with WH-ACE, the loss function of ExCTC training is
\begin{align}
L_{ExCTC} = L_{CTC} + L_{WH-ACE}.
\end{align}

IFA-inference applies $\mathbb{C}$ and $\mathbb{F}$ to the input image, \ie, $\bm{Y}=\mathbb{C}(\mathbb{F}(\bm{x}))$. 
After merging the eight-neighborhoods of $argmax(Y)$, we can get the recognition result $Y'$. $Y'$ is a set of $(x, y, c)$, where $x$, $y$, $c$ represent the horizontal coordinate, vertical coordinate and category of the character, respectively. 
To translate the recognition results into text sequences, we design a simple rule-based postprocessing \ie, Algorithm.~\ref{algorithm_decoding}, which follows the priori that documents can be read from left to right and from top to bottom.
$\lambda_{x}$ and $\lambda_{y}$ are hyper-parameters, and are set as $20$ and $2$ in this work. 

The implementation of IFA is rather simple that it can be realized by adding WH-ACE on attention-based methods or on CTC-based methods with column squeeze module.

\begin{algorithm}[h] 
\caption{Postprocessing} 
\begin{algorithmic}
\label{algorithm_decoding} 
\REQUIRE ~~\\
Recognition result $Y'$
\ENSURE  ~~\\
Text sequence $O$
\STATE $Y' \leftarrow sort(Y')$ sorted in ascending order of $x$
\STATE $C' \leftarrow \varnothing, O \leftarrow \varnothing, I \leftarrow Lengthof\{Y'\}$
\WHILE{$Y' != \varnothing$}
\STATE $C \leftarrow \{c_{0}\}, x_{p} \leftarrow x_{0}, y_{p} \leftarrow y_{0}$
\FOR{$i \in [1, 2, ..., I]$}
\IF{$x_{i} - x_{p} < \lambda_{x}, abs(y_{i} -y_{p}) < \lambda_{y}$}
\STATE $x_{p} \leftarrow x_{i}, y_{p} \leftarrow y_{i}, C \leftarrow C \cup \{c_{i}\}, Y' \leftarrow Y' - (x_{i}, y_{i}, c_{i})$
\ENDIF 
\ENDFOR
\STATE $C' \leftarrow C' \cup \{(y_{0}, C)\}, I \leftarrow Lengthof\{Y'\}$
\ENDWHILE
\STATE $C' \leftarrow sort(C')$ sorted in ascending order of $y$
\FOR{$i \in [1, 2, ..., Lengthof\{C'\}]$}
\STATE $O \leftarrow O \cup C_{i}$
\ENDFOR
\end{algorithmic} 
\end{algorithm}

\section{Experiments}
\subsection{Datasets and Implementation Details}

\subsubsection{Datasets}

\textbf{IAM} \cite{marti2002iam} dataset is based on handwritten English text copied from the LOB corpus. It contains 747 documents (6,482 lines) in the training set, 116 documents (976 lines) in the validation set and 336 documents (2,915 lines) in the testing set.

\textbf{CASIA-HWDB} \cite{liu2011casia} is a large-scale Chinese handwritten database. In this study, we use two offline datasets: CASIA-HWDB 1.0-1.2 and CASIA-HWDB 2.0-2.2. CASIA-HWDB 1.0-1.2 contains 3,895,135 isolated character samples. CASIA-HWDB 2.0-2.2 contains 5,091 pages.

\textbf{ICDAR-2013} \cite{yin2013icdar} includes a page-level dataset. There are 300 pages in ICDAR-2013. This dataset is used to evaluate the models trained on CASIA-HWDB.

\textbf{Touching-Block} is an in-house Chinese educational handwritten dataset with 1,358 images (4,810 lines). This dataset encounters serious touching problem, where characters are densely and closely arranged, as shown in Fig.~\ref{Figure_Touching}. Experimental models on this dataset are trained on CASIA-HWDB as well as another in-house training set with about 100k lines.

\textbf{OffRaSHME} \cite{OffRaSHME} is a handwritten mathematical recognition datasets with 19,749 training images. This dataset is the largest public mathematical recognition dataset reported to date.

\subsubsection{Implementation Details}

On the IAM and ICDAR-2013 dataset, the height of the training image is normalized to 80 pix and the width is calculated with the original aspect ratio (up to 1200 pix).
During inference on IAM testing set and CASIA dataset, the multi-line image is scaled 0.5$\times$ and then input into the model.
On OffRaSHME dataset, the height of the training image is normalized to 128 pix and the width is calculated with the original aspect ratio (up to 768 pix).
We adopt ResNet-18 \cite{he2016deep} as backbone in all experiments.
In IAM, the evaluation criteria is the character error rate (CER\%), corresponding to the edit distance between the recognition result and ground-truth, normalized by the length of ground-truth.
In ICDAR-2013 and Touching-block, the evaluation criteria is correct rate (CR) and accuracy rate (AR) specified by ICDAR2013 competition\cite{yin2013icdar}.

\begin{table}[b]
\small
\caption{Ablation study of the use of WH-ACE. The measure of performance is the CER(\%).}
\label{Table_ablation_whace}
\begin{center}
\begin{tabular}{|c|c|c|c|}
\hline
\textbf{Methods}& \textbf{WH-ACE} & \textbf{Conventional} & \textbf{IFA-inference}\\ \hline
\multirow{2}{*}{\textbf{ADP}} & & 11.85 & 27.65\\ \cline{3-4}
& \checkmark & 11.78 & 10.28 \\ \hline
\multirow{2}{*}{\textbf{ExCTC}} & & 6.48 & 15.83 \\ \cline{3-4}
& \checkmark & 6.42 & 6.26 \\ \hline
\end{tabular}
\end{center}
\end{table}

\begin{figure}[b]
\centering
\includegraphics[width=0.4\textwidth]{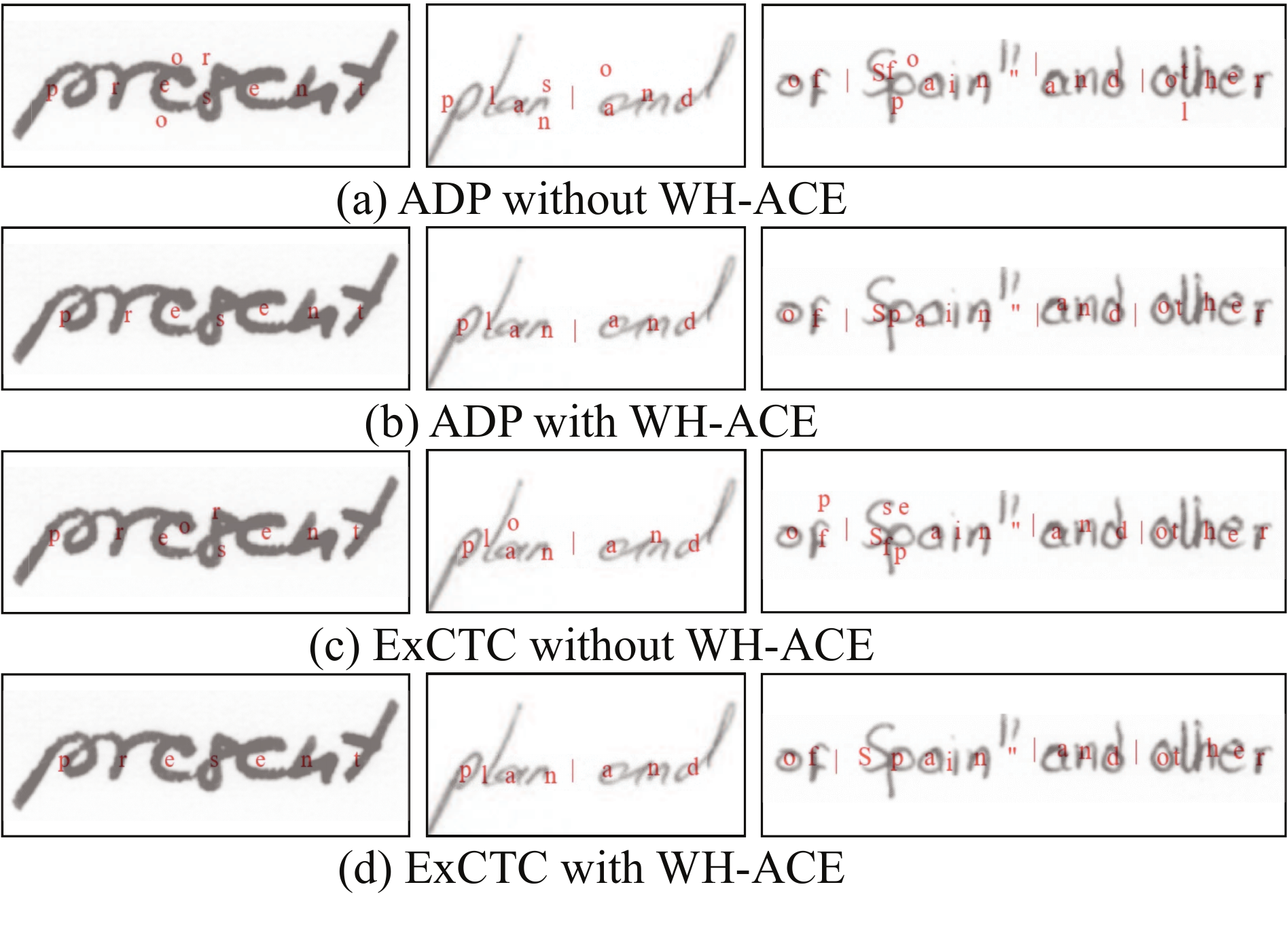}
\caption{Visualization of the use of WH-ACE.}
\label{Figure_ACEVS}
\end{figure}

\subsection{Effectiveness Validation of WH-ACE}

In this subsection, we verify the noise-suppression ability of WH-ACE, which is used to complement ADP and ExCTC for negative sample suppression.

Here, we compare conventional inference and IFA-inference of ADP and ExCTC. 
It can be observed from Table. \ref{Table_ablation_whace} that the performance of conventional inference is almost the same with or without WH-ACE, while the performance of IFA-inference has an evident improvement when training with WH-ACE.
This experimental result is consistent with our expectation, because we expect WH-ACE to suppress the negative noises without affecting the training of positive pixels.

A few visualization results are shown in Fig.~\ref{Figure_ACEVS}.
Without WH-ACE, some strokes on the corners of characters are easily recognized as similar characters; for example, the upper part of ``p" is recognized as ``o" in Fig.~\ref{Figure_ACEVS} (a).

\begin{table}[b]
\footnotesize
\caption{Performance comparison from single-line to multi-line text recognition. The measure of performance is the CER(\%). `IFA-line' and `IFA-fullpage' denotes IFA-inference works on line text and full-page text.}
\label{Table_C2U}
\begin{center}
\begin{tabular}{|c|c|c|c|c|c|}
\hline
\textbf{Methods}& \textbf{Conventional} & \textbf{IFA-line} & \textbf{IFA-fullpage}& \textbf{$g_1$}& \textbf{$g_2$} \\ \hline
ExCTC & 6.42 & 6.26 & 6.16 & 0.16 & 0.10 \\ \hline
ADP & 11.78 & 10.28 & 9.97 & 1.50 & 0.31 \\ \hline
\end{tabular}
\end{center}
\end{table}

\begin{figure}[b]
\centering
\includegraphics[width=0.42\textwidth]{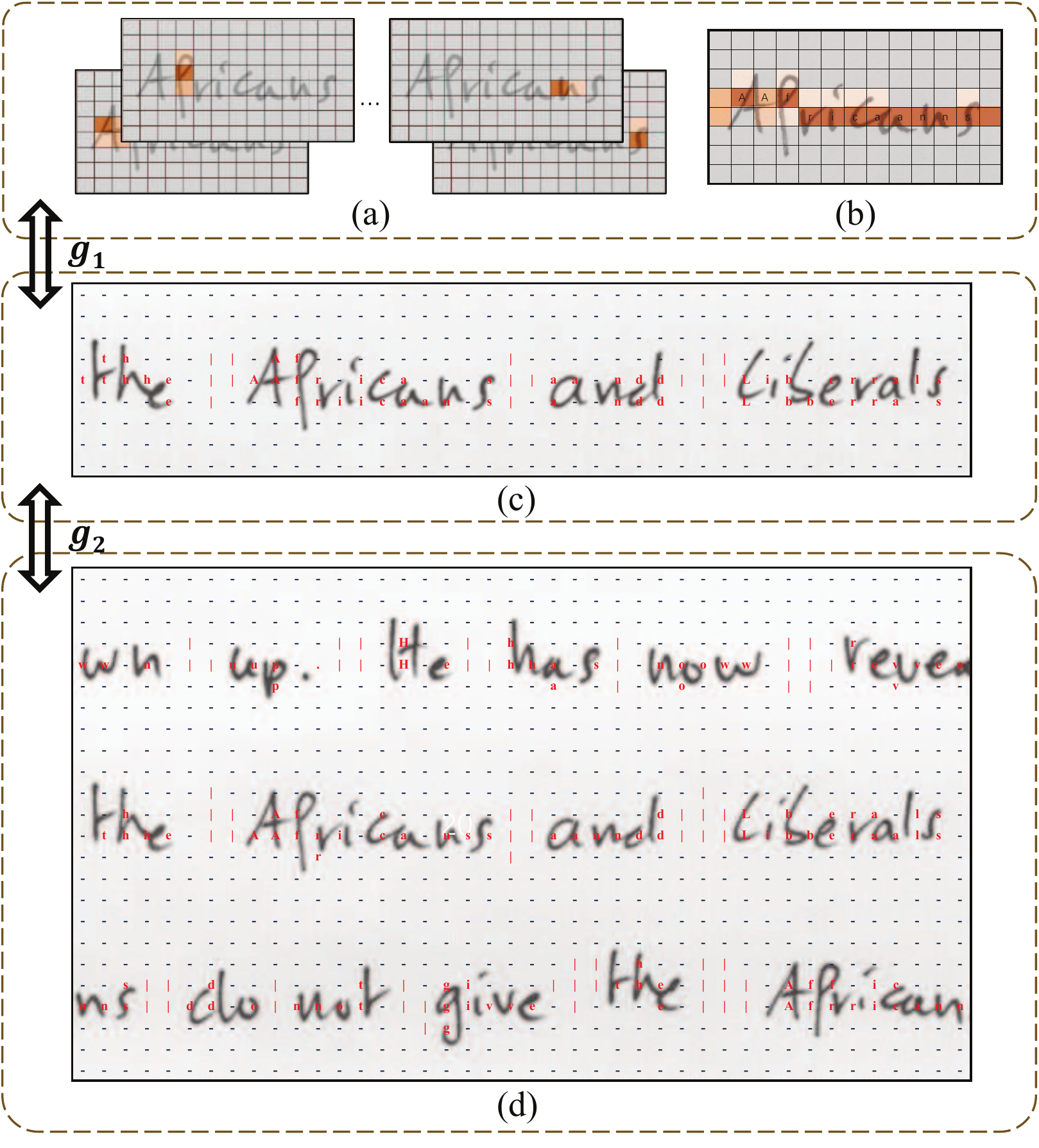}
\caption{The conversion from line text recognition to multi-line recognition. (a) conventional-inference of ADP on single-line. (b) conventional-inference of ExCTC on single-line. (c) IFA-inference on single-line. (d) IFA-inference on multi-line.}
\label{Figure_GAP}
\end{figure}

\subsection{Convert Text Recognizer to Text Spotter}
In this subsection, we will quantitatively analyze the process of converting single-line recognizer to multi-line recognizer by IFA.

The conversion can be divided into two stages, the first stage is the conversion from conventional inference to IFA-inference on single lines, and the second stage is the conversion of IFA-inference from single-line to multi-line images.
To facilitate the analyze, we define two gaps, $g_1$ and $g_2$, corresponding to the performance degradation of these two stages, which are visually shown in Fig.~\ref{Figure_GAP}.

The results are given in Table. \ref{Table_C2U}. For ExCTC, $g_1$ is close to 0, which implies that the conversion brought by IFA involves no performance degradation.
For ADP, $g_1$ is larger than 1\%, implying that the inference form after conversion is even better than that of the original attention decoder. 
This is because attention decoder often encounters alignment problem on long sequence \cite{cong2019comparative}, 
and the conversion replaces the attention decoder with a more robust one-step classification operation on each pixels.
For both ExCTC and ADP, $g_2$ is also close to 0, which implies that IFA can effectively process both single-line and multi-line texts.

In summary, IFA-inference enables a conventional text recognizer to process multi-line texts, serving as a detection-free text spotter. 

\begin{table}[b]
\small
\caption{Comparison with other state-of-the-art methods on IAM dataset. We adopt the ResNet version of OrigamiNet \cite{yousef2020origaminet} for fair comparison.}
\label{Table_IAM}
\begin{center}
\begin{tabular}{|c|c|c|c|}
\hline
\textbf{Methods}& \textbf{Year} & \textbf{CER} \\ \hline
Bluche \cite{bluche2016joint} & 2016 & 7.9 \\
SFR \cite{2018Start} & 2018 & 6.4 \\
E2E HTR \cite{carbonell2019end} & 2019 & 15.6 \\
OrigamiNet \cite{yousef2020origaminet} & 2020 & \textbf{6.1} \\ \hline \hline

ADP & 2020 & 9.97 \\
ExCTC & 2020 & 6.16 \\ \hline
\end{tabular}
\end{center}
\end{table}

\subsection{IFA vs Previous Methods: Advantages}
In this subsection, we will compare IFA with the traditional Detection-Recognition (Det-Recog) OCR systems on Latin (IAM) and non-Latin (ICDAR-2013 and touching-block) datasets, and we demonstrate that IFA is faster and more robust.
On the IAM dataset, we conduct a comparison with other published results.
On the ICDAR-2013 and touching-block datasets, we reproduce several popular methods and compare the performances of ExCTC with them.

\begin{figure}[t]
\centering
\includegraphics[width=0.48\textwidth]{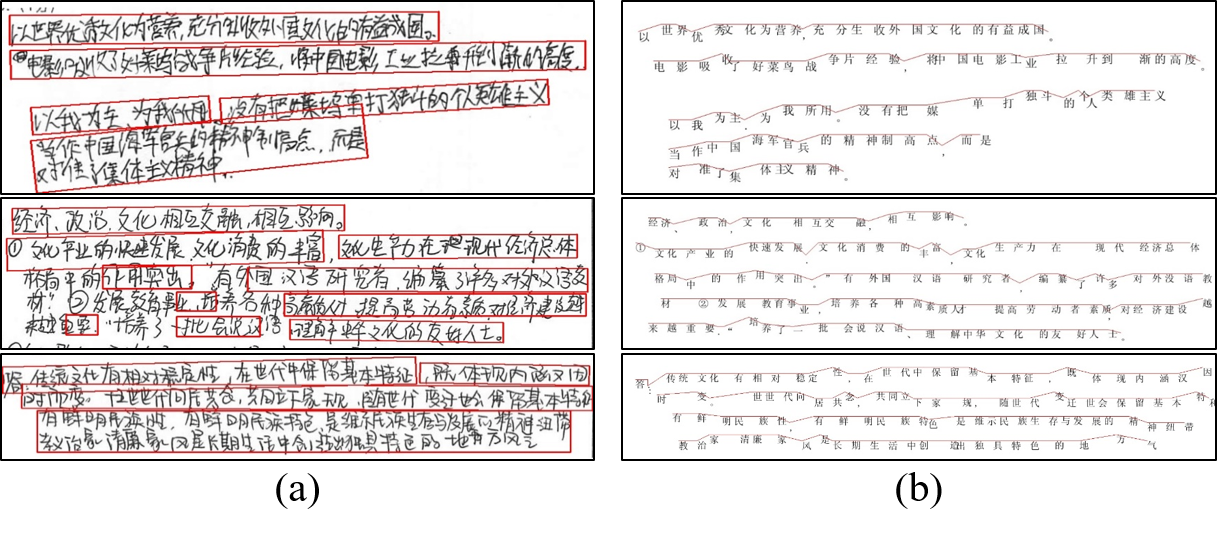}
\caption{Visualization of Det-Recog and ExCTC on touching-block dataset. (a) Visualization of Det-Recog pipeline, red boxes denote detection results. (b) Visualization of ExCTC, red line links the characters that belong to the same line text via Algorithm.~\ref{algorithm_decoding}.}
\label{Figure_Touching}
\end{figure}

For Latin text recognition, as shown in Table. \ref{Table_IAM}, ExCTC surpass all Det-Recog methods and achieves competitive performance with the current state-of-the-art OrigamiNet.
For non-Latin text recognition, as shown in Table. \ref{Table_ICDAR}, ExCTC achieves state-of-the-art performance with the fastest speed.
It surpasses the traditional Det-Recog method, methods designed for scene text spotting (MaskText Spotter \cite{lyu2018mask} and FOTS \cite{liu2018fots}) and methods designed for document full-page OCR (SFR \cite{2018Start} and OrigamiNet \cite{yousef2020origaminet}).
It has a speed of 11.4 times higher than Det-Recog methods, 19.1 times faster than MaskText Spotter \cite{lyu2018mask} and 1.1 times faster than FOTS \cite{liu2018fots}.
On the touching-block dataset, IFA appears to be much better than Det-Recog (AR 84.96\% \emph{vs} AR 77.75\%). This is because IFA directly ignore the detection process which is error-prone in this serious touching dataset. 
Although OrigamiNet \cite{yousef2020origaminet} performs well in Latin text recognition, its performance degrades significantly in non-Latin text recognition, whereas ExCTC maintains high performance in both Latin and non-Latin text recognition.
Some visualization results on touching-block dataset are shown in Fig.~\ref{Figure_Touching}.

\begin{table}[t]
\small
\caption{Comparison with other state-of-the-art methods on ICDAR dataset. The `Det + Recog' is the combination of two independently trained models which are Faster RCNN \cite{ren2015faster} equipped with RRPN \cite{ma2018arbitrary} for line text detection and a recognizer\cite{shi2016end} for line text recognition. Speed is tested with NVIDIA GTX 1080ti.}
\label{Table_ICDAR}
\begin{center}
\begin{tabular}{|c|c|c|c|}
\hline
\textbf{Methods}& \textbf{AR} & \textbf{CR} & \textbf{Speed(fps)} \\ \hline
Det + Recog & 89.65 & 90.39 & 2.79 \\ \hline
Mask Textspotter \cite{lyu2018mask} & 50.60 & 58.29 & 1.61 \\ \hline
FOTS \cite{liu2018fots} & 67.32 & 67.82 & 27.03 \\ \hline
SFR \cite{2018Start} & 82.91 & 83.55 & 0.61 \\ \hline
OrigamiNet \cite{yousef2020origaminet} & 5.99 & 5.99 & 3.19 \\ \hline \hline
ExCTC & \textbf{89.97} & \textbf{90.56} & \textbf{31.76} \\ \hline
\end{tabular}
\end{center}
\end{table}

\subsection{ADP vs ExCTC: application scenarios}
The above experiments results show that ExCTC has better performance than ADP on document full-page recognition task. 
There are two reasons for this phenomenon.
First, CTC-based methods have been proved to be more reliable than attention-based methods on sentence recognition tasks \cite{cong2019comparative}.
And their derivative methods inherited this property.
Second, ADP cannot separate adjacent characters of the same class very well, but ExCTC can avoid this error by inserting a blank symbol.
This error may occur when facing words like ``well", ``apple".
However, attention is more flexible than CTC when facing complex 2-D text; for example, mathematical expressions.

On OffRaSHME dataset, we use the longest 4,000 samples for validation and the rest for training.
It may be noted that we do not aim to fully recognize these expressions, but to verify the wide range of applications of ADP. Thus, we only consider the character recognition rate.

It can be observed in Table 5 that ADP has a better character recognition rate than these attention-based methods.
It surpasses the state-of-the-art WAP-TD \cite{zhang2020treedecoder} by 4\%.
This is because ADP leverages the flexible alignment property of the attention mechanism while abandoning the complex and uncontrollable step-by-step decoding process, which is error-prone on long sequences.

In summary, ADP and ExCTC both have their own application scenarios. 
ADP can handle more complex recognition problems, while ExCTC is more effective for regular document recognition. 
They have their own advantages and complement each other in different application scenarios.

\begin{table}[t]
\small
\caption{Performance comparison on OffRaSHME dataset.}
\label{Table_crohme}
\begin{center}
\begin{tabular}{|c|c|c|c|}
\hline
\textbf{Methods}& \textbf{Precision} & \textbf{Recall} & \textbf{F-measure} \\ \hline
WAP \cite{zhang2017watch} & 89.49 & 78.79 & 83.80 \\ \hline
WAP + TD \cite{zhang2020treedecoder} & 93.43 & 77.48 & 84.74\\ \hline \hline
ExCTC & 29.96 & 10.21 & 15.23 \\ \hline
ADP & \textbf{94.00} & \textbf{85.08} & \textbf{89.32} \\ \hline
\end{tabular}
\end{center}
\end{table}

\section{Discussion}

Compared with previous studies, in this study, IFA first unifies the form of text recognition and text spotting. 
IFA-inference can work on single-line and multi-line images, resulting in a much simpler OCR system.

Although IFA can directly conduct detection-free text spotting, the current version still requires rule-based post-processing to generate text from dense predictions, which has low generality. 
In the future, we will explore a better linking strategy to replace the current rule-based post-processing and extend the method to scene-text spotting tasks.

\section{Conclusion}
In this paper, we proposes a simple, yet effective, paradigm called IFA to convert a text recognizer into a detection-free text spotter.
IFA leverages the learnable alignment property of neural network and can be easily integrated into current mainstream text recognizers. 
Specifically, we integrate IFA with the attention mechanism and CTC, resulting in two new methods: ADP and ExCTC, respectively.
In addition, we propose WH-ACE to suppress the negative noise. 
Through comprehensive experiments, we find that IFA maintains state-of-the-art performance on several document recognition tasks with a significantly simpler pipeline, and both ADP and ExCTC have their own merits in different application scenarios.

\section*{Acknowledgement}

This research is supported in part by NSFC (Grant No.: 61936003, 61771199),  GD-NSF (no.2017A030312006), and the National Key Research and Development Program  of China (No. 2016YFB1001405).



{\small
\bibliographystyle{ieee_fullname}
\bibliography{egbib}

\begin{thebibliography}{10}\itemsep=-1pt

\bibitem{baek2020character}
Youngmin Baek, Seung Shin, Jeonghun Baek, Sungrae Park, Junyeop Lee, Daehyun
  Nam, and Hwalsuk Lee.
\newblock Character region attention for text spotting.
\newblock In {\em Eur. Conf. Comput. Vis.}, pages 504--521. Springer, 2020.

\bibitem{bahdanau2015neural}
Dzmitry Bahdanau, Kyunghyun Cho, and Yoshua Bengio.
\newblock Neural machine translation by jointly learning to align and
  translate.
\newblock In {\em Int. Conf. Learn. Represent.}, 2015.

\bibitem{bluche2016joint}
Th{\'e}odore Bluche.
\newblock Joint line segmentation and transcription for end-to-end handwritten
  paragraph recognition.
\newblock In {\em Adv. Neural Inform. Process. Syst.}, pages 838--846, 2016.

\bibitem{2017Deep}
Michal Busta, Lukas Neumann, and Jiri Matas.
\newblock Deep textspotter: An end-to-end trainable scene text localization and
  recognition framework.
\newblock In {\em Int. Conf. Comput. Vis.}, pages 2223--2231, 2017.

\bibitem{carbonell2019end}
Manuel Carbonell, Joan Mas, Mauricio Villegas, Alicia Forn{\'e}s, and Josep
  Llad{\'o}s.
\newblock End-to-end handwritten text detection and transcription in full
  pages.
\newblock In {\em International Conference on Document Analysis and Recognition
  Workshops (ICDAR-W)}, volume~5, pages 29--34. IEEE, 2019.

\bibitem{cong2019comparative}
Fuze Cong, Wenping Hu, Qiang Huo, and Li Guo.
\newblock A comparative study of attention-based encoder-decoder approaches to
  natural scene text recognition.
\newblock In {\em International Conference on Document Analysis and Recognition
  (ICDAR)}, pages 916--921. IEEE, 2019.

\bibitem{feng2019textdragon}
Wei Feng, Wenhao He, Fei Yin, Xu-Yao Zhang, and Cheng-Lin Liu.
\newblock Textdragon: An end-to-end framework for arbitrary shaped text
  spotting.
\newblock In {\em Int. Conf. Comput. Vis.}, pages 9076--9085, 2019.

\bibitem{Graves2006Connectionist}
Alex Graves, Santiago Fernández, Faustino~J. Gomez, and Jürgen Schmidhuber.
\newblock Connectionist temporal classification: labelling unsegmented sequence
  data with recurrent neural networks.
\newblock In {\em International Conference on Machine Learning}, pages
  369--376, 2006.

\bibitem{graves2009novel}
Alex Graves, Marcus Liwicki, Santiago Fern{\'a}ndez, Roman Bertolami, Horst
  Bunke, and J{\"u}rgen Schmidhuber.
\newblock A novel connectionist system for unconstrained handwriting
  recognition.
\newblock {\em IEEE Trans. Pattern Anal. Mach. Intell.}, 31(5):855--868, 2009.

\bibitem{gupta2016synthetic}
Ankush Gupta, Andrea Vedaldi, and Andrew Zisserman.
\newblock Synthetic data for text localisation in natural images.
\newblock In {\em IEEE Conf. Comput. Vis. Pattern Recog.}, pages 2315--2324,
  2016.

\bibitem{he2016deep}
Kaiming He, Xiangyu Zhang, Shaoqing Ren, and Jian Sun.
\newblock Deep residual learning for image recognition.
\newblock In {\em IEEE Conf. Comput. Vis. Pattern Recog.}, pages 770--778,
  2016.

\bibitem{huang2019attention}
Yunlong Huang, Canjie Luo, Lianwen Jin, Qingxiang Lin, and Weiying Zhou.
\newblock Attention after attention: Reading text in the wild with cross
  attention.
\newblock In {\em International Conference on Document Analysis and Recognition
  (ICDAR)}, pages 274--280. IEEE, 2019.

\bibitem{jaderberg2014synthetic}
Max Jaderberg, Karen Simonyan, Andrea Vedaldi, and Andrew Zisserman.
\newblock Synthetic data and artificial neural networks for natural scene text
  recognition.
\newblock In {\em Proceedings of Advances in Neural Information Processing Deep
  Learn. Workshop (NIPS-W)}, 2014.

\bibitem{lee2016recursive}
Chen-Yu Lee and Simon Osindero.
\newblock Recursive recurrent nets with attention modeling for ocr in the wild.
\newblock In {\em IEEE Conf. Comput. Vis. Pattern Recog.}, pages 2231--2239,
  2016.

\bibitem{li2017towards}
Hui Li, Peng Wang, and Chunhua Shen.
\newblock Towards end-to-end text spotting with convolutional recurrent neural
  networks.
\newblock In {\em Int. Conf. Comput. Vis.}, pages 5238--5246, 2017.

\bibitem{li2019show}
Hui Li, Peng Wang, Chunhua Shen, and Guyu Zhang.
\newblock Show, attend and read: A simple and strong baseline for irregular
  text recognition.
\newblock In {\em AAAI}, pages 8610--8617, 2019.

\bibitem{liao2019scene}
Minghui Liao, Jian Zhang, Zhaoyi Wan, Fengming Xie, Jiajun Liang, Pengyuan Lyu,
  Cong Yao, and Xiang Bai.
\newblock Scene text recognition from two-dimensional perspective.
\newblock In {\em AAAI}, pages 8714--8721, 2019.

\bibitem{litman2020scatter}
Ron Litman, Oron Anschel, Shahar Tsiper, Roee Litman, Shai Mazor, and R
  Manmatha.
\newblock Scatter: selective context attentional scene text recognizer.
\newblock In {\em IEEE Conf. Comput. Vis. Pattern Recog.}, pages 11962--11972,
  2020.

\bibitem{liu2011casia}
Cheng-Lin Liu, Fei Yin, Da-Han Wang, and Qiu-Feng Wang.
\newblock Casia online and offline chinese handwriting databases.
\newblock {\em International Conference on Document Analysis and Recognition
  (ICDAR)}, pages 37--41, 2011.

\bibitem{liu2018char}
Wei Liu, Chaofeng Chen, and Kwan-Yee~K Wong.
\newblock {Char-Net}: A character-aware neural network for distorted scene text
  recognition.
\newblock In {\em AAAI}, pages 7154--7161, 2018.

\bibitem{liu2018fots}
Xuebo Liu, Ding Liang, Shi Yan, Dagui Chen, Yu Qiao, and Junjie Yan.
\newblock Fots: Fast oriented text spotting with a unified network.
\newblock In {\em IEEE Conf. Comput. Vis. Pattern Recog.}, pages 5676--5685,
  2018.

\bibitem{liu2020abcnet}
Yuliang Liu, Hao Chen, Chunhua Shen, Tong He, Lianwen Jin, and Liangwei Wang.
\newblock Abcnet: Real-time scene text spotting with adaptive bezier-curve
  network.
\newblock In {\em IEEE Conf. Comput. Vis. Pattern Recog.}, pages 9809--9818,
  2020.

\bibitem{long2020unrealtext}
Shangbang Long and Cong Yao.
\newblock Unrealtext: Synthesizing realistic scene text images from the unreal
  world.
\newblock In {\em IEEE Conf. Comput. Vis. Pattern Recog.}, pages 5488--5497,
  2020.

\bibitem{cluo2019moran}
Canjie Luo, Lianwen Jin, and Zenghui Sun.
\newblock {MORAN}: A multi-object rectified attention network for scene text
  recognition.
\newblock {\em Pattern Recognition}, 90:109--118, 2019.

\bibitem{luong2015effective}
Thang Luong, Hieu Pham, and Christopher~D Manning.
\newblock Effective approaches to attention-based neural machine translation.
\newblock In {\em Conference on Empirical Methods in Natural Language
  Processing}, pages 1412--1421, 2015.

\bibitem{lyu2018mask}
Pengyuan Lyu, Minghui Liao, Cong Yao, Wenhao Wu, and Xiang Bai.
\newblock Mask textspotter: An end-to-end trainable neural network for spotting
  text with arbitrary shapes.
\newblock In {\em Eur. Conf. Comput. Vis.}, pages 67--83, 2018.

\bibitem{ma2018arbitrary}
Jianqi Ma, Weiyuan Shao, Hao Ye, Li Wang, Hong Wang, Yingbin Zheng, and
  Xiangyang Xue.
\newblock Arbitrary-oriented scene text detection via rotation proposals.
\newblock {\em IEEE Trans. Multimedia}, 20(11):3111--3122, 2018.

\bibitem{marti2002iam}
U-V Marti and Horst Bunke.
\newblock The iam-database: an english sentence database for offline
  handwriting recognition.
\newblock {\em International Journal on Document Analysis and Recognition},
  5(1):39--46, 2002.

\bibitem{qiao2019text}
Liang Qiao, Sanli Tang, Zhanzhan Cheng, Yunlu Xu, Yi Niu, Shiliang Pu, and Fei
  Wu.
\newblock Text perceptron: Towards end-to-end arbitrary-shaped text spotting.
\newblock In {\em AAAI}, 2020.

\bibitem{qiao2020seed}
Zhi Qiao, Yu Zhou, Dongbao Yang, Yucan Zhou, and Weiping Wang.
\newblock Seed: Semantics enhanced encoder-decoder framework for scene text
  recognition.
\newblock In {\em IEEE Conf. Comput. Vis. Pattern Recog.}, pages 13528--13537,
  2020.

\bibitem{qin2019towards}
Siyang Qin, Alessandro Bissacco, Michalis Raptis, Yasuhisa Fujii, and Ying
  Xiao.
\newblock Towards unconstrained end-to-end text spotting.
\newblock In {\em Int. Conf. Comput. Vis.}, pages 4704--4714, 2019.

\bibitem{ren2015faster}
Shaoqing Ren, Kaiming He, Ross Girshick, and Jian Sun.
\newblock Faster r-cnn: Towards real-time object detection with region proposal
  networks.
\newblock In {\em Adv. Neural Inform. Process. Syst.}, pages 91--99, 2015.

\bibitem{shi2016end}
Baoguang Shi, Xiang Bai, and Cong Yao.
\newblock An end-to-end trainable neural network for image-based sequence
  recognition and its application to scene text recognition.
\newblock {\em IEEE Trans. Pattern Anal. Mach. Intell.}, 39(11):2298--2304,
  2016.

\bibitem{shi2016robust}
Baoguang Shi, Xinggang Wang, Pengyuan Lyu, Cong Yao, and Xiang Bai.
\newblock Robust scene text recognition with automatic rectification.
\newblock In {\em IEEE Conf. Comput. Vis. Pattern Recog.}, pages 4168--4176,
  2016.

\bibitem{shi2018aster}
Baoguang Shi, Mingkun Yang, Xinggang Wang, Pengyuan Lyu, Cong Yao, and Xiang
  Bai.
\newblock {ASTER}: An attentional scene text recognizer with flexible
  rectification.
\newblock {\em IEEE Trans. Pattern Anal. Mach. Intell.}, 2018.

\bibitem{wan2020textscanner}
Zhaoyi Wan, Minghang He, Haoran Chen, Xiang Bai, and Cong Yao.
\newblock Textscanner: Reading characters in order for robust scene text
  recognition.
\newblock In {\em AAAI}, volume~34, pages 12120--12127, 2020.

\bibitem{OffRaSHME}
Da-Han Wang, Fei Yin, Jin-Wen Wu, Yu-Pei Yan, Zhi-Cai Huang, Gui-Yun Chen, Yao
  Wang, and Cheng lin Liu.
\newblock {ICFHR} 2020 competition on offline recognition and spotting of
  handwritten mathematical expressions.
\newblock \url{http://offrashme.xmutpr.com/home}.

\bibitem{wang2020all}
Hao Wang, Pu Lu, Hui Zhang, Mingkun Yang, Xiang Bai, Yongchao Xu, Mengchao He,
  Yongpan Wang, and Wenyu Liu.
\newblock All you need is boundary: Toward arbitrary-shaped text spotting.
\newblock In {\em AAAI}, volume~34, pages 12160--12167, 2020.

\bibitem{wang2012handwritten}
Qiu-Feng Wang, Fei Yin, and Cheng-Lin Liu.
\newblock Handwritten chinese text recognition by integrating multiple
  contexts.
\newblock {\em IEEE Trans. Pattern Anal. Mach. Intell.}, 34(8):1469--1481,
  2012.

\bibitem{wang2019scene}
Siwei Wang, Yongtao Wang, Xiaoran Qin, Qijie Zhao, and Zhi Tang.
\newblock Scene text recognition via gated cascade attention.
\newblock In {\em Int. Conf. Multimedia and Expo}, pages 1018--1023, 2019.

\bibitem{DAN_aaai20}
Tianwei Wang, Yuanzhi Zhu, Lianwen Jin, Canjie Luo, Xiaoxue Chen, Yaqiang Wu,
  Qianying Wang, and Mingxiang Cai.
\newblock Decoupled attention network for text recognition.
\newblock In {\em AAAI}, 2020.

\bibitem{2018Start}
Curtis Wigington, Chris Tensmeyer, Brian Davis, William Barrett, and Scott
  Cohen.
\newblock Start, follow, read: End-to-end full-page handwriting recognition.
\newblock In {\em Eur. Conf. Comput. Vis.}, 2018.

\bibitem{wu2017improving}
Yi-Chao Wu, Fei Yin, and Cheng-Lin Liu.
\newblock Improving handwritten chinese text recognition using neural network
  language models and convolutional neural network shape models.
\newblock {\em Pattern Recognition}, 65:251--264, 2017.

\bibitem{Xie2019Aggregation}
Zecheng Xie, Yaoxiong Huang, Yuanzhi Zhu, Lianwen Jin, and Lele Xie.
\newblock Aggregation cross-entropy for sequence recognition.
\newblock In {\em IEEE Conf. Comput. Vis. Pattern Recog.}

\bibitem{yang2019symmetry}
Mingkun Yang, Yushuo Guan, Minghui Liao, Xin He, Kaigui Bian, Song Bai, Cong
  Yao, and Xiang Bai.
\newblock Symmetry-constrained rectification network for scene text
  recognition.
\newblock In {\em Int. Conf. Comput. Vis.}, pages 9147--9156, 2019.

\bibitem{yang2017learning}
Xiao Yang, Dafang He, Zihan Zhou, Daniel Kifer, and C~Lee Giles.
\newblock Learning to read irregular text with attention mechanisms.
\newblock In {\em IJCAI}, pages 3280--3286, 2017.

\bibitem{yin2013icdar}
Fei Yin, Qiu-Feng Wang, Xu-Yao Zhang, and Cheng-Lin Liu.
\newblock {ICDAR} 2013 chinese handwriting recognition competition.
\newblock {\em International Conference on Document Analysis and Recognition
  (ICDAR)}, pages 1464--1470, 2013.

\bibitem{yousef2020origaminet}
Mohamed Yousef and Tom~E Bishop.
\newblock Origaminet: Weakly-supervised, segmentation-free, one-step, full page
  text recognition by learning to unfold.
\newblock In {\em IEEE Conf. Comput. Vis. Pattern Recog.}, pages 14710--14719,
  2020.

\bibitem{Deli2020Towards}
Deli Yu, Xuan Li, Chengquan Zhang, Junyu Han, Jingtuo Liu, and Errui Ding.
\newblock Towards accurate scene text recognition with semantic reasoning
  networks.
\newblock In {\em IEEE Conf. Comput. Vis. Pattern Recog.}, pages 12113--12122,
  2020.

\bibitem{zhan2019esir}
Fangneng Zhan and Shijian Lu.
\newblock {ESIR}: End-to-end scene text recognition via iterative image
  rectification.
\newblock In {\em IEEE Conf. Comput. Vis. Pattern Recog.}, pages 2059--2068,
  2019.

\bibitem{zhan2018verisimilar}
Fangneng Zhan, Shijian Lu, and Chuhui Xue.
\newblock Verisimilar image synthesis for accurate detection and recognition of
  texts in scenes.
\newblock In {\em Eur. Conf. Comput. Vis.}, pages 249--266, 2018.

\bibitem{zhan2019ga}
Fangneng Zhan, Chuhui Xue, and Shijian Lu.
\newblock Ga-dan: Geometry-aware domain adaptation network for scene text
  detection and recognition.
\newblock In {\em Int. Conf. Comput. Vis.}, pages 9105--9115, 2019.

\bibitem{zhang2020treedecoder}
Jianshu Zhang, Jun Du, Yongxin Yang, Yi-Zhe Song, Si Wei, and Lirong Dai.
\newblock A tree-structured decoder for image-to-markup generation.
\newblock In {\em International Conference on Machine Learning}, page In Press,
  2020.

\bibitem{zhang2017watch}
Jianshu Zhang, Jun Du, Shiliang Zhang, Dan Liu, Yulong Hu, Jinshui Hu, Si Wei,
  and Lirong Dai.
\newblock Watch, attend and parse: An end-to-end neural network based approach
  to handwritten mathematical expression recognition.
\newblock {\em Pattern Recognition}, 71:196--206, 2017.

\bibitem{zhou2013handwritten}
Xiang-Dong Zhou, Da-Han Wang, Feng Tian, Cheng-Lin Liu, and Masaki Nakagawa.
\newblock Handwritten chinese/japanese text recognition using semi-markov
  conditional random fields.
\newblock {\em IEEE Trans. Pattern Anal. Mach. Intell.}, 35(10):2413--2426,
  2013.

\bibitem{zhou2014minimum}
Xiang-Dong Zhou, Yan-Ming Zhang, Feng Tian, Hong-An Wang, and Cheng-Lin Liu.
\newblock Minimum-risk training for semi-markov conditional random fields with
  application to handwritten chinese/japanese text recognition.
\newblock {\em Pattern Recognition}, 47(5):1904--1916, 2014.

\bibitem{zhu2019text}
Yiwei Zhu, Shilin Wang, Zheng Huang, and Kai Chen.
\newblock Text recognition in images based on transformer with hierarchical
  attention.
\newblock In {\em IEEE Int. Conf. Image Process.}, pages 1945--1949, 2019.

\end{thebibliography}
}

\end{document}